\documentclass[a4paper,american,reqno]{amsart}

\usepackage[utf8]{inputenc}
\usepackage[T1]{fontenc}
\usepackage{babel}
\usepackage[binary-units=true]{siunitx}
\usepackage[ruled,linesnumbered]{algorithm2e}
\usepackage{csquotes}
\usepackage{todonotes}
\usepackage{multicol}
\usepackage{comment}
\usepackage{booktabs}
\usepackage{flafter}
\usepackage{longtable}
\usepackage{dsfont}
\usepackage{placeins}
\usepackage{float}
\allowdisplaybreaks
\usepackage[style = authoryear-comp,
            maxbibnames = 100,
            maxcitenames = 2,
            giveninits = true,
            uniquename = init,
            isbn = false,
            backend = bibtex]{biblatex}
\usepackage[colorlinks,
            citecolor=blue,
            urlcolor=blue,
            linkcolor=blue]{hyperref} 

\makeatletter
\patchcmd{\@settitle}{\uppercasenonmath\@title}{\scshape\large}{}{}
\patchcmd{\@setauthors}{\MakeUppercase}{\scshape\normalsize}{}{}
\makeatother

\vfuzz \hfuzz
\makeatletter
\@namedef{subjclassname@2020}{%
  \textup{2020} Mathematics Subject Classification}
\makeatother

\SetKwInOut{Input}{Input}
\SetKwInOut{Output}{Output}



\newcommand\MyBox[2]{
  \fbox{\lower0.75cm
    \vbox to 1.7cm{\vfil
      \hbox to 1.7cm{\hfil\parbox{1.4cm}{#1\\#2}\hfil}
      \vfil}%
  }%
}

\bibliography{main}

\begin{document}

\title[Fair Mixed Effects Support Vector Machine]%
{Fair Mixed Effects Support Vector Machine}

\author[J. P. Burgard, J. V. Pamplona]%
{Jan Pablo Burgard, João Vitor Pamplona}

\address[J. P. Burgard, J. V. Pamplona]{%
  Trier University,
  Department of Economic and Social Statistics,
  Universitätsring 15,
  54296 Trier,
  Germany}
\email{burgardj@uni-trier.de, pamplona@uni-trier.de}

\date{\today}

\begin{abstract}
To ensure unbiased and ethical automated classifications, fairness must be a core principle in machine learning applications. Fairness in machine learning aims to mitigate biases present in the training data and model imperfections that could lead to discriminatory outcomes. This is achieved by preventing the model from making decisions based on sensitive characteristics like ethnicity or sexual orientation. A fundamental assumption in machine learning is the independence of observations. However, this assumption often does not hold true for data describing social phenomena, where data points are often clustered based. Hence, if the machine learning models do not account for the cluster correlations, the results may be biased. Especially high is the bias in cases where the cluster assignment is correlated to the variable of interest. We present a fair mixed effects support vector machine algorithm that can handle both problems simultaneously. With a reproducible simulation study we demonstrate the impact of clustered data on the quality of fair machine learning classifications.
\end{abstract}

\keywords{Support Vector Machine, 
Fair Machine Learning,
Mixed Models.
}
\subjclass[2020]{90C90,
90-08,
68T99
%
%
%
}

\maketitle

\section{Introduction}
\label{sec:introduction}

The rise of automated decision-making systems calls for the development of fair algorithms. These algorithms must be constrained by societal values, particularly to avoid discrimination against any population group. While machine learning offers efficiency gains, it can also unintentionally perpetuate biases in critical areas like loan approvals \parencite{das2021fairness} and criminal justice \parencite{green2018fair}. In loan applications, factors like marital status can lead to unfair disadvantages for single individuals. At the same time, in criminal justice, algorithms might associate race with recidivism risk, leading to discriminatory sentencing despite individual circumstances. This highlights the need for fair and unbiased AI frameworks to ensure equal opportunities and outcomes for all.

Driven by the need to mitigate bias in algorithms, fair machine learning is experiencing a surge in research activity. Numerous research articles are exploring innovative approaches to achieve fairer outcomes. Notable examples include fair versions of Logistic and Linear Regression \parencite{fairRL2}, Support Vector Machines (SVM) \parencite{FairSVM2}, Random Forests \parencite{fairRF}, Decision Trees \parencite{FairDT}, and Generalized Linear Models (GLMs) \parencite{fairGLM}. These methods aim to address potential discrimination arising from historical data or algorithmic design, ensuring fairer outcomes for all individuals.

A key challenge in machine learning for automated decision-making is the training data. Often sourced from surveys, this data may not perfectly align with the assumption common in machine learning that all elements are sampled independently with equal probability of inclusion. However, some data may suffer from cluster effects, e.g., in marketing, customers who buy a particular product might also be interested in a similar product. Ignoring these effects can lead to misleading results, such as underestimating the true variability in the data. To overcome it, random effects are incorporated into the statistical model, which, together with the existing fixed effects, leads to a mixed effects model, as can be seen in \textcite{oberg2007linear}.

In this paper we propose a Mixed Effects Support Vector Machine for fair classifications. We show how to estimate the model and evaluate its performance against the standard SVM model that does not take the possible clustering of the data into account. Similar approaches for longitudinal data can be found in \textcite{SVM1} and \textcite{SVM3}, for Least-squares support vector machine in \textcite{SVM5} and applications in agricultural activities in \textcite{SVM2}.

The paper is organized as follows: In Section \ref{sec:chapter-2} we explore the theory and metrics behind fairness in machine learning and how it applies to support vector machines.
In Section \ref{sec:chapter-3} we establish the theoretical underpinnings of fair mixed effects support vector machine and propose a strategy for solving it.
In Section \ref{sec:chapter-4}, we conduct a comprehensive evaluation of our proposed method's effectiveness through various tests. Finally, in Section \ref{sec:chapter5}, we demonstrate the practical applicability of the algorithm by solving a real-world problem using the Adult dataset \parencite{Adult}.
Our key findings and potential future directions are presented in Section \ref{sec:conclusion}.

\section{Fair Support Vector Machine}
\label{sec:chapter-2}

Classification algorithms in machine learning are used to estimate a specific classification $\hat{y}\in \{-1,1\}$ for a new data point $x$ based on a training set $\mathcal{D} = {(x^{\ell}, y_{\ell})}^n_{\ell=1}$. For the point $x^\ell$, if $y_\ell = 1$, we say that $x^\ell$ is in the positive class and if $y_\ell=-1$, $x^\ell$ belongs to the negative class with  $x^{\ell} \in \mathbb{R}^{p+1}$, for each $\ell \in [1,n] := \{1, \cdots, n\}$, belonging to data $X = \left[x^1, \dotsc, x^n\right]$ being the dimension of $x$, $p+1$, due to the addition of an extra column with the value $1$ as the data intercept.

In the realm of fair binary classification, each observation $x^{\ell}$ possesses a corresponding sensitive attribute $s_\ell$ a binary value of either $0$ or $1$, the goal becomes identifying a solution that balances accuracy with fairness. Fairness in machine learning can be assessed using various metrics. In this paper,  we specifically focus on disparate impact (DI).  Alternative fairness metrics for binary classifiers can be found in \textcite{bigdata}.

Consider  $\mathcal{S}_1 = \{x^\ell : s_\ell = 1\}$ and $\mathcal{S}_0 = \{x^\ell : s_\ell = 0\}$ as disjoint subsets of dataset $X$, where the sensitive feature of all points in this subset is $1$ and $0$, respectively. A binary classifier is considered free of disparate impact if the probability of classifying an instance as either $0$ or $1$ is equal in $\mathcal{S}_1$ and $\mathcal{S}_0$. In other terms, disparate impact is absent when the classifier treats all groups equally regardless of their sensitive feature values. Mathematically, we can write this as:
\[
P(\hat{y}_{\ell} = 1 | x_\ell \in \mathcal{S}_0 ) = P(\hat{y}_{\ell} = 1 |x_\ell \in \mathcal{S}_1 ).
\]

To this end, we must maintain a proportional relationship between the classifications for both classes of the sensitive feature. Defining 
\[
 \dfrac{\displaystyle\sum_{\ell=1}^n s_\ell}{n} =  \dfrac{\vert \mathcal{S}_1 \vert}{n} =: \Bar{s} 
\] 
because $s_\ell \in \{0, 1\}$ and knowing that in SVM, proposed by \textcite{vapnik1964class}, a hyperplane $\beta^\top x$ splits the feature space of the data based on the classification of each point we want to maintain the proportionality of classifications for both sensitive categories, so:
\begin{equation*}
    \frac{\vert \mathcal{S}_0 \vert}{n}\displaystyle\sum_{x^\ell \in \mathcal{S}_1} \beta^\top x^{\ell} = \frac{\vert \mathcal{S}_1 \vert}{n}\displaystyle\sum_{x^\ell \in \mathcal{S}_0} \beta^\top x^{\ell}.
\end{equation*}
This means that
\begin{align*}
    &1 - \frac{\vert \mathcal{S}_1 \vert}{n}\displaystyle\sum_{x^\ell \in \mathcal{S}_1} \beta^\top x^{\ell} = \frac{\vert \mathcal{S}_1 \vert}{n}\displaystyle\sum_{x^\ell \in \mathcal{S}_0} \beta^\top x^{\ell}\\
    &\implies (1 - \Bar{s})\displaystyle\sum_{x^\ell \in \mathcal{S}_1} \beta^\top x^{\ell} = \Bar{s}\displaystyle\sum_{x^\ell \in \mathcal{S}_0} \beta^\top x^{\ell} \\
    &\implies \displaystyle\sum_{x^\ell \in \mathcal{S}_1} (1 - \Bar{s})\beta^\top x^{\ell} = \displaystyle\sum_{x^\ell \in \mathcal{S}_0} (\Bar{s}-0)\beta^\top x^{\ell} \\
    &\implies \displaystyle\sum_{x^\ell \in \mathcal{S}_1} (1 - \Bar{s})\beta^\top x^{\ell} + \displaystyle\sum_{x^\ell \in \mathcal{S}_0} (0-\Bar{s})\beta^\top x^{\ell} = 0 \\
    &\implies \displaystyle\sum_{\ell=1}^{n} \displaystyle (s_{\ell} - \bar{s})(\beta^\top x^{\ell}) = 0.
\end{align*}
To maintain consistency with the previously proposed literature \parencite{pmlr-v54-zafar17a}, we divide the sum above by $n$. Note that this does not alter the equality.

In the following, we present how to add fairness in the context of Support Vector Machine (SVM). Note that in our formulation the first column of X is equal to 1 for all points and, hence, the bias parameter is the first entry in $\beta$ \parencite{hsieh2008dual}.
By adding the fairness constraint, the fixed effect $\beta$ can be found by solving the following quadratic optimization problem.

\begin{subequations}\label{SVMDI}
\begin{align}
\underset {\left(\beta, \xi\right)}{\min} &\;\;\;\;
\frac{1}{2} \Vert \beta \Vert^2 + K \sum_{\ell=1}^N \xi_{\ell}\label{ret2222}\\ 
\rm{s.t} &\;\;\;\; y_{\ell}(\beta^\top x_{\ell} ) \geq 1 - \xi_{\ell}, \text{  } \ell = 1, \dots, N\label{ret22223}, \\
&\;\;\;\; \frac{1}{N}\sum\limits_{\ell=1}^N(s_{\ell}-\Bar{s})(\beta^{\top}x_{\ell}) \leq c\label{ret22221},\\
&\;\;\;\; \frac{1}{N}\sum\limits_{\ell=1}^N(s_{\ell}-\Bar{s})(\beta^{\top}x_{\ell}) \geq -c\label{ret22222}, \\ &\;\;\;\; \xi_{\ell} \geq 0, \text{  } \ell = 1, \dots, N. \label{ret2222233}
\end{align}
\end{subequations}

Constraints \eqref{ret2222}, \eqref{ret22223} and \eqref{ret2222233} are from the Support Vector Machine problem. Constraints \eqref{ret22221} and \eqref{ret22222}  guarantees the fairness in the classification. This model is proposed by \textcite{Paper1}. While achieving zero disparate impact is a desirable goal, it can potentially come at the expense of a good classification as we have a trade-off between fairness and accuracy \parencite{tradeoff1,tradeoff2}. To address this exchange, we can introduce a fairness threshold, denoted by $c \in \mathbb{R}^+$, which allows us to adjust the relative importance placed on fairness compared to accuracy. The penalty parameter $K$ aims to control the importance of slack variables, which represents the flexibility in classifying a point considering the optimal hyperplane. We refer to the problem above as Fair Support Vector Machine (SVMF).

In the following example, the two sensitive categories are distinguished by the shape of the point (diamond and star). The color differentiates the label, with red being positive $(1)$ and blue being negative $(-1)$.

\begin{figure}[!ht]
    \centering
    \begin{minipage}[t]{0.5\textwidth}
        \centering
        \includegraphics[width=1\textwidth]{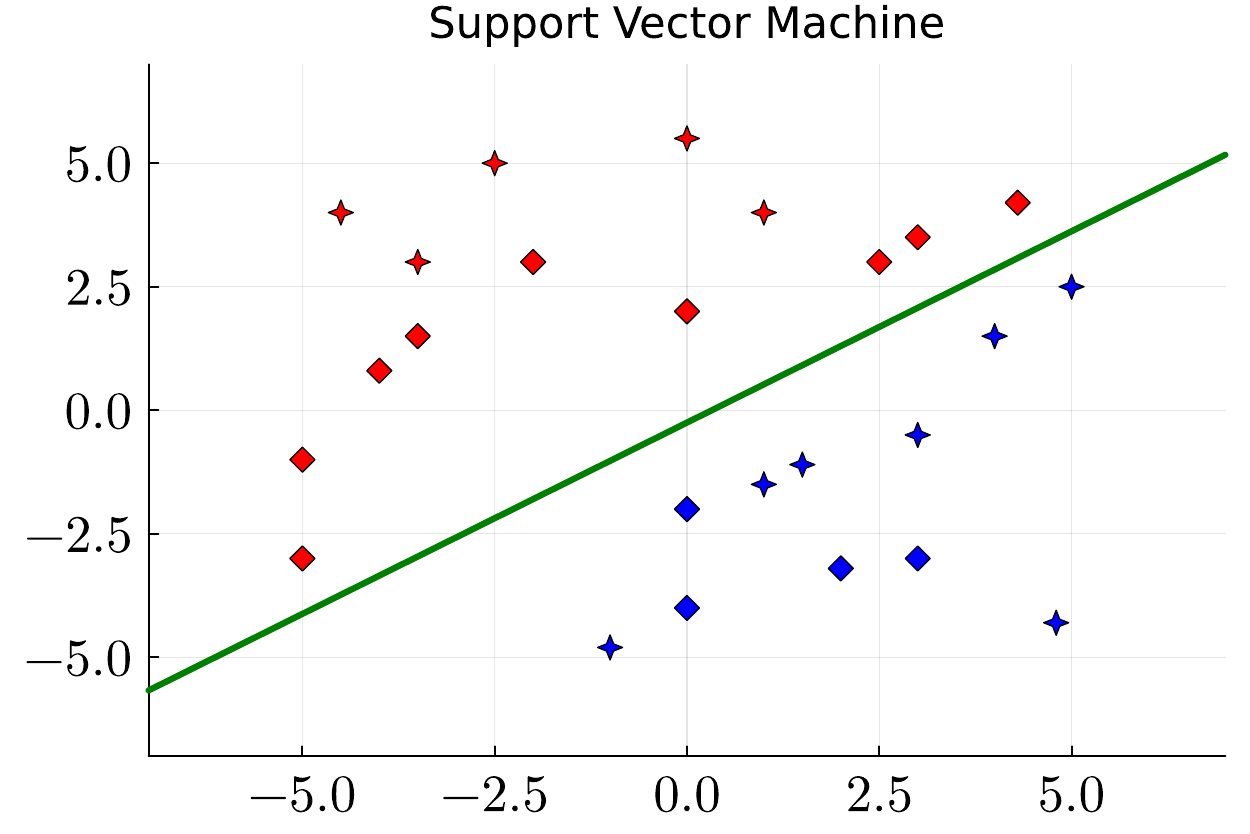} 
        \caption{Regular SVM.}\label{fig1}
    \end{minipage}\hfill
    \begin{minipage}[t]{0.5\textwidth}
        \centering
        \includegraphics[width=1\textwidth]{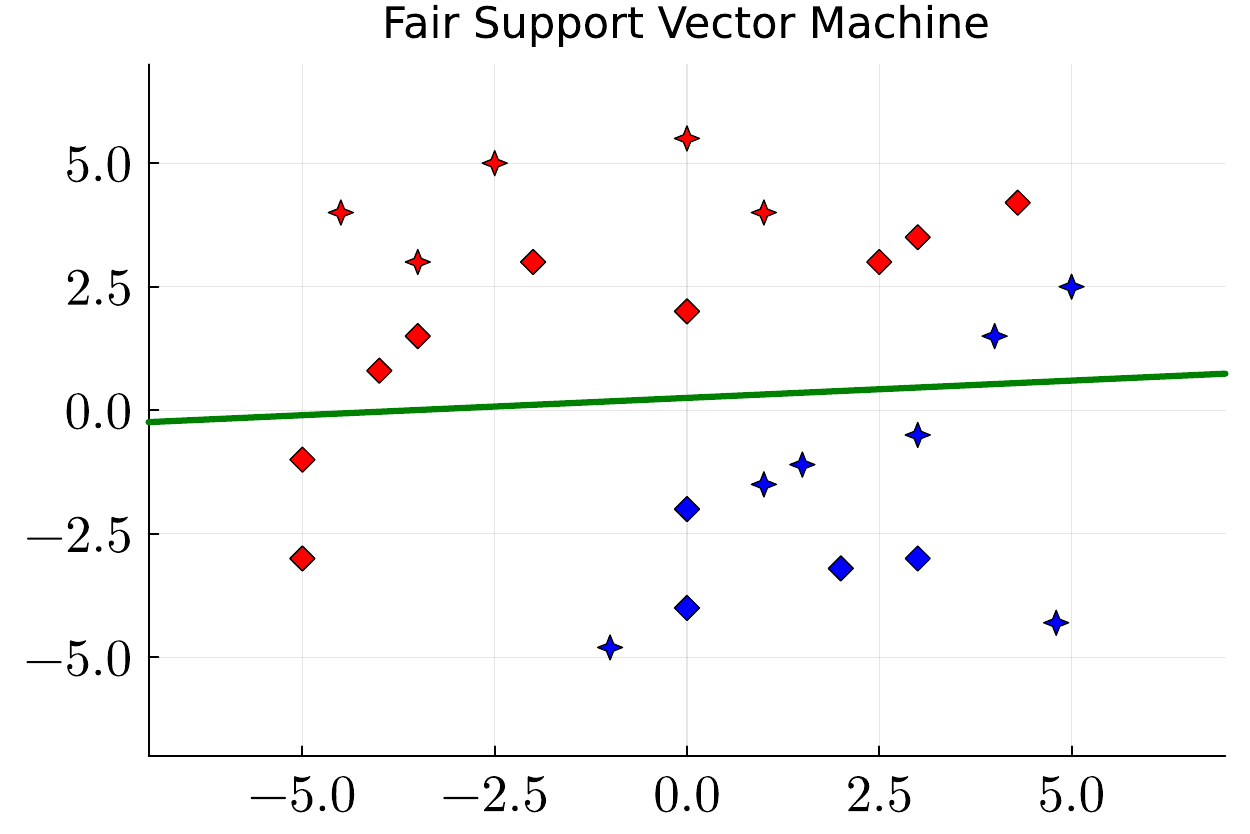}
        \caption{SVM free of disparate impact.}\label{fig2}
    \end{minipage}
\end{figure}

Figure \ref{fig1} demonstrates $100\%$ accuracy, as each point is located on the correct side of the hyperplane. However, when we look at each sensitive feature category, we observe that approximately $36\% (5/14)$ of the star points are classified as positive while this happens for approximately $64\% (9/14)$ of the diamond points. That is, the proportions of the two sensitive categories being classified as positive are not equal.

Figure \ref{fig2} shows an example of SVM free of disparate impact as in Problem \eqref{SVMDI}, considering $c=0$. We can see that $50\%$ of the star points and approximately $53\%$ of the diamond points are classified as positive. That is, the proportion of positive classifications is similar for both sensitive categories. However this comes at the cost of accuracy, as $4$ points were misclassified, reducing the accuracy to $84\%$. This can be mitigated by choosing a threshold $c$, sufficiently good so as not to significantly impair accuracy.

\section{Fair Mixed Effects Support Vector Machine}
\label{sec:chapter-3}


Real-world data often shows heterogenic variations within groups. Consider a dataset comprising multiple schools where we aim to ascertain whether distinct teachers yield varying outcomes for a particular variable. In a predictive modeling context, we would classify these teaching capacity variables as fixed effects, that is effects that represents factors that are of direct interest and are explicitly modeled \parencite{agasisti2017heterogeneity}. However, it is plausible that other unmeasured factors, such as the quality of teaching materials or the overall school environment, also influence the outcomes. These unmeasured factors are typically modeled as random effects, that is factors that are not of primary interest but introduce variability into the data \parencite{greene1997econometric}. 

To account for the heterogeneity introduced by these random effects and to obtain unbiased estimates of the impact of teachers on student outcomes, it is imperative to incorporate them into the predictive model. Neglecting these random effects could lead to substantial bias in the classifications, particularly when comparing outcomes across different schools. Statistical tools can estimate their impact on grouped data.

Let $g$ being the random vector and $g_i$ with $i ~\in~ [1,K]$, representing the group-specific random effect, with $g$ following a normal distribution with mean zero. Consider $\Gamma_i$ the size of the group $i$ for each $i ~\in~ [1,K]$ and $y_{ij}$ the label of $(x^{ij})^\top~=~(x^{ij}_1, \dots, x^{ij}_p)$ with $j \in [1,\Gamma_i]$.

In essence, the fair constraints discussed in the section above only consider fixed effects ($\beta$'s). However, in mixed models we need to consider the random effects $g_i$, that is, $\beta^{\top} x^{ij} + g_i$ being the inner product used to compute the probability of the point $x^{ij}$ being classified as $1$. In light of these considerations, we propose the following adaptation in the constraints:
\begin{multicols}{2}
\noindent
\[
\mathcal{S}_1^i = \{x^{ij} : j \in [1,\Gamma_i], s_{ij} = 1\}
\]

\noindent
\[
\mathcal{S}_0^i = \{x^{ij} : j \in [1,\Gamma_i], s_{ij} = 0\}
\]
\end{multicols}
Observe that each subset is created for each cluster $i \in [1,K]$.

Moreover, we need to modify the prediction function in the fair constraints to account for random effects.

Following the same logic as presented before, but considering a group-to-group analysis, we have a similar construction for the disparate impact constraints:

\noindent\begin{align*}
    &\frac{\vert \mathcal{S}_0 \vert}{n}\displaystyle\sum_{i=1}^K\displaystyle\sum_{x^{ij} \in \mathcal{S}_1^i} (\beta^\top x^{ij} + g_i) = \frac{\vert \mathcal{S}_1 \vert}{n}\displaystyle\sum_{i=1}^K\displaystyle\sum_{x^{ij} \in \mathcal{S}_0^i} (\beta^\top x^{ij} + g_i) \\ &\implies 1 - \frac{\vert \mathcal{S}_1 \vert}{n}\displaystyle\sum_{i=1}^K\displaystyle\sum_{x^{ij} \in \mathcal{S}_1^i} (\beta^\top x^{ij} + g_i) = \frac{\vert \mathcal{S}_1 \vert}{n}\displaystyle\sum_{i=1}^K\displaystyle\sum_{x^{ij} \in \mathcal{S}_0^i} (\beta^\top x^{ij} + g_i)\\
    &\implies (1 - \Bar{s})\displaystyle\sum_{i=1}^K\displaystyle\sum_{x^{ij} \in \mathcal{S}_1^i} (\beta^\top x^{ij} + g_i) = \Bar{s}\displaystyle\sum_{i=1}^K\displaystyle\sum_{x^{ij} \in \mathcal{S}_0^i} (\beta^\top x^{ij} + g_i) \\
    &\implies \displaystyle\sum_{i=1}^K\displaystyle\sum_{x^{ij} \in \mathcal{S}_1^i} (1 - \Bar{s})(\beta^\top x^{ij} + g_i) = \displaystyle\sum_{i=1}^K\displaystyle\sum_{x^{ij} \in \mathcal{S}_0^i} (\Bar{s}-0)(\beta^\top x^{ij} + g_i) \\
    &\implies \displaystyle\sum_{i=1}^K\displaystyle\sum_{x^{ij} \in \mathcal{S}_1^i} (1 - \Bar{s})(\beta^\top x^{ij} + g_i) - \displaystyle\sum_{i=1}^K\displaystyle\sum_{x^{ij} \in \mathcal{S}_0^i} (0-\Bar{s})(\beta^\top x^{ij} + g_i) = 0 \\
    &\implies \displaystyle\sum_{i=1}^K \displaystyle\sum_{j=1}^{\Gamma_i} (s_{ij} + \bar{s})(\beta^\top x^{ij} + g_i) = 0.
\end{align*}

Besides that is important control the random effects variance we do that penalizing it through L2 regularization, aiming to minimize it. Similar approaches for least-square support vector machine can be found in \textcite{SVM1} and as demonstrated in \textcite{domingos2012few,zou2005regularization,friedman2010regularization}, minimizing the variance enhances model performance and generalization. High variance often implies that the model is excessively sensitive to the training data, resulting in sub-optimal performance on unseen data. Hence the sum $\sum_i^K g_i^2$ is from the fact that the variance of $g$ is given by:
\[
\sigma^2 = \frac{\sum_{i=1}^K(g_i - \Bar{g})^2}{K-1},
\]
and the random effect $g$ follows a normal distribution with mean zero. 
\begin{align*}
    \sigma^2 &= \frac{\sum_{i=1}^K(g_i - 0)^2}{K-1}
    = \frac{\sum_{i=1}^K g_i^2}{K-1}.
\end{align*}

This means that the regularization term is $\frac{\sum_{i=1}^K g_i^2}{K-1}$. Since we have a minimization problem and $K-1$ is fixed, we can consider w.l.o.g a parameter $\lambda$ that controls the importance of the random effects.

Consequently, we formulate the following constrained optimization problem:

\begin{subequations}\label{MESVMDI}
\begin{align}
\underset {\left(\beta, b  ,\xi \right )}{\min} &\;\;\;\;
\frac{1}{2} \Vert \beta \Vert^2 + \mu \sum_{i=1}^K \sum_{j=1}^{\Gamma_i} \xi_{ij} + \lambda \sum_{i=1}^K g_i^2\label{f1}\\ 
\rm{s.t} &\;\;\;\; y_{ij}(m_{\beta,g}^{SVM}(x^{ij})) \geq 1 - \xi_{ij},\label{f2} \\
 &\;\;\;\; \frac{1}{n}\sum\limits_{i=1}^{K}\sum\limits_{j=1}^{\Gamma_i}(s_{ij}-\Bar{s})(\beta^\top x^{ij} + g_i)  \leq c\label{f3},\\
\; &\;\;\;\; \frac{1}{n}\sum\limits_{i=1}^{K}\sum\limits_{j=1}^{\Gamma_i}(s_{ij}-\Bar{s})(\beta^\top x^{ij} + g_i) \geq -c\label{f4},\\
&\;\;\;\; \xi_{ij} \geq 0, \text{  } i = 1, \dots, K, \text{  } j = 1, \dots, \Gamma_i,\label{f5}
\end{align}
\end{subequations}
The objective function \eqref{f1} and the constraints \eqref{f2} and \eqref{f5} are a adapted version of Support Vector Machine problem to deal with random effects. Constraints \eqref{f3} and \eqref{f4}  guarantees the fairness, in terms of disparate impact. We refer to the problem above as Fair Mixed Effects Support Vector Machine (FMESVM) and without the fairness constraints as Mixed Effects Support Vector Machine (MESVM).

In our numerical experiments, we set variables $\mu$ and $\lambda$ equal to $1$. The choice of $\mu$ reflects the desire to assign equal importance to maximizing the margin and minimizing the classification error. The value of $\lambda$ is determined by the aim of balancing the influence of random and fixed effects within the optimization problem. Preliminary tests revealed that setting $\lambda$ significantly higher than $1$ tends to prioritize random effects, resulting in a random vector with excessive variance, even when the objective is to minimize it. On the other hand, if $\lambda$ is considerably lower than $1$, the optimization process becomes overly focused on fixed effects, leading to random effects that are almost negligible.

\subsection*{A One-hot encoding alternative}
One alternative approach to incorporating group effects into grouped data analysis involves the use of dummy variables, also known as one-hot encoding as can be seen in \textcite{inbook234}. This strategy involves creating a separate binary variable for each group, taking a value of $1$ for observations within that group and $0$ otherwise.  To the best of our knowledge, no one has previously employed this technique in the SVM setting in the literature; however, we consider this method, and, while it can effectively capture group-level variation, preliminary numerical results have demonstrated that the proposed approach offers significant advantages in terms of both computational efficiency and memory usage, as can be seen below. This happens because,  in the one-hot encoding, the dimension of each $x^\ell$ is increased in the number $K$ of groups, that is,  $x\in \mathbb{R}^{p+1+K}$. 

The following preliminary numerical experiment employed a dataset consisting of $100000$ points with the training set having 3 to 5 points per group and systematically varying the number of groups within the data, considering configurations with $2$, $10$, $50$, $500$, $1000$, $1250$, $2000$, $2500$, $3125$, $4000$, and $5000$ groups.

\begin{figure}[H]
\centering
\includegraphics[width=12.5cm]{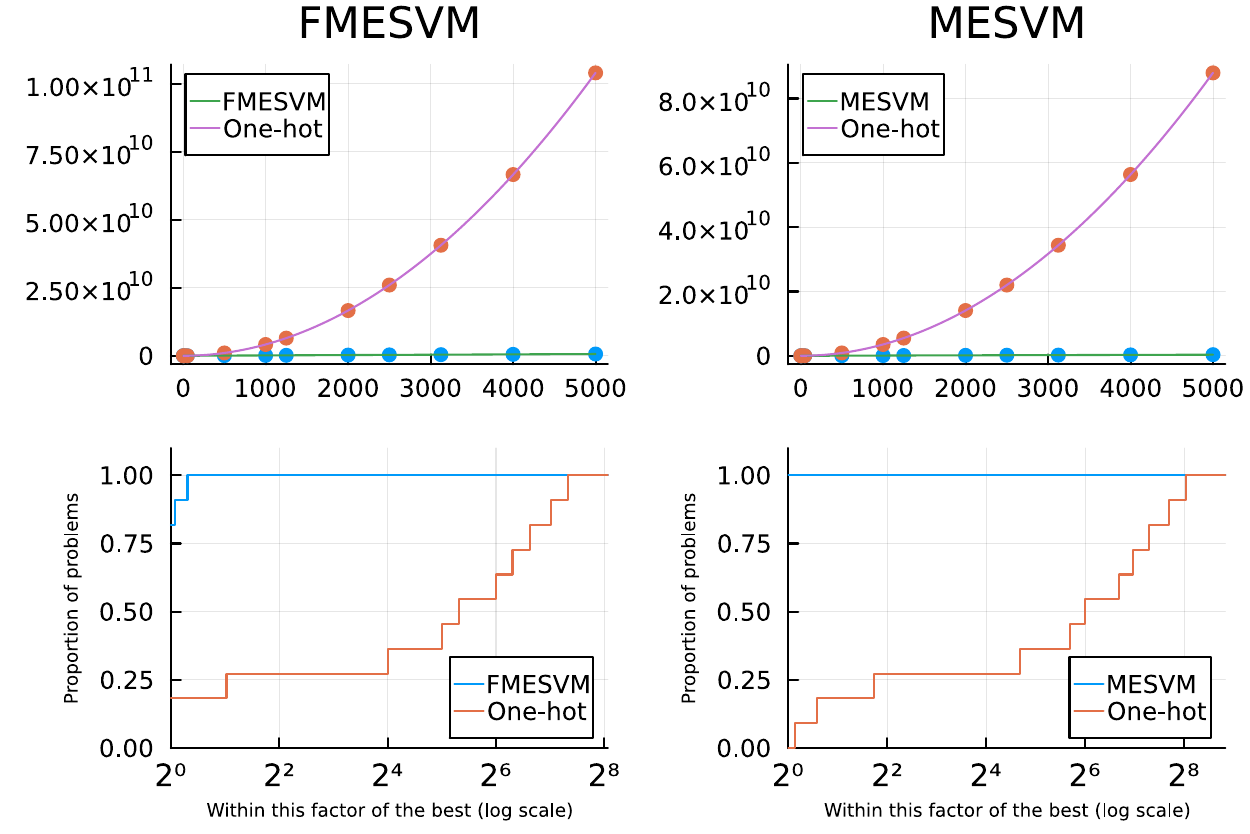}
\caption{Memory comparison in Bytes.}
\label{MMBY}
\end{figure}

\begin{figure}[H]
\centering
\includegraphics[width=12.5cm]{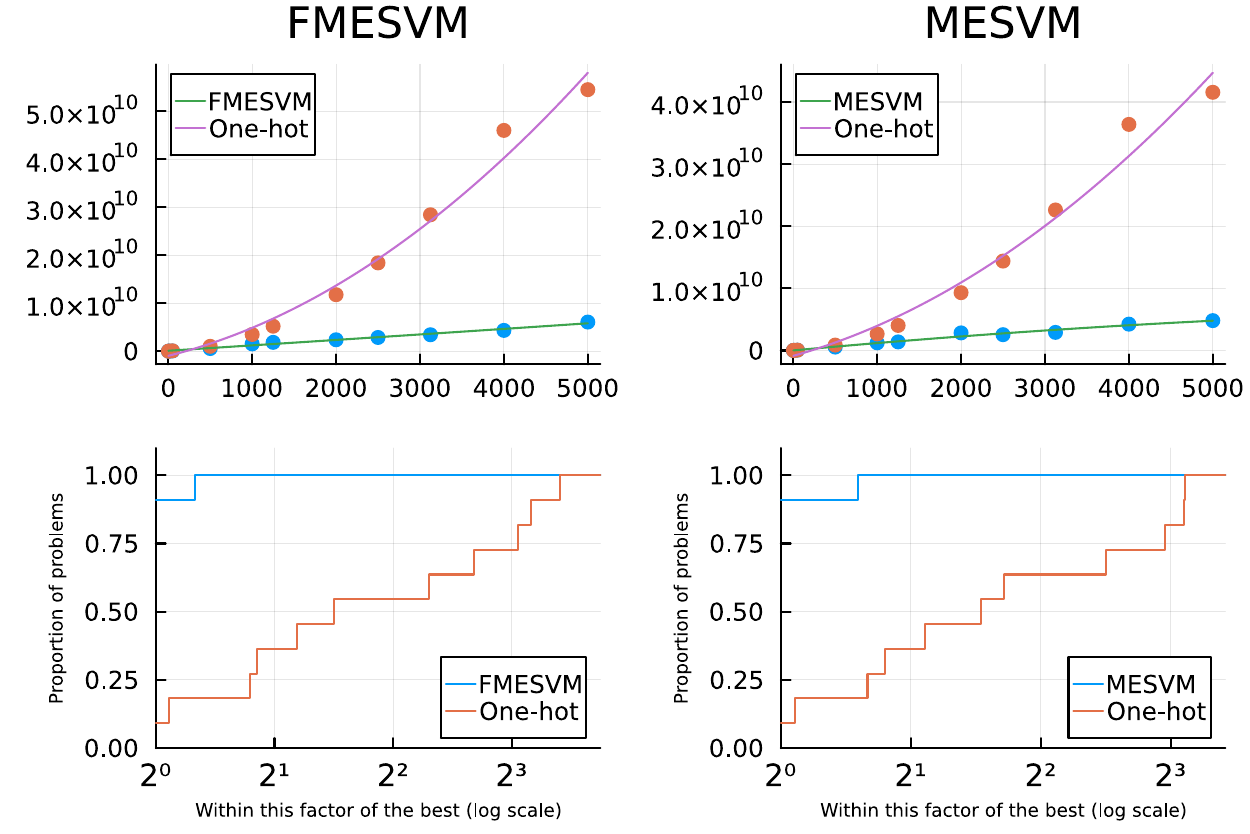}
\caption{Time comparison in Microsecond.}
\label{TTMS}
\end{figure}

Figures \ref{MMBY} shows the Memory comparison in Bytes and Figure \ref{TTMS} the time comparison. The first row of each figure present a second-order polynomial fit, where the x-axis represents the number of groups and the y-axis the memory. The second row present the Performance Profile proposed by \textcite{DolanMore02} of both approaches. In this specific numerical test, the problem \eqref{MESVMDI} is the Mixed Effects Support Vector Machine free of Disparate Impact and is labeled as FMESVM.

As can be seen in Figures \ref{MMBY} and \ref{TTMS}, our optimization models outperforms the one-hot encoding in time and memory. Hence, the approach is a more resource-friendly option, mainly for large datasets and complex models in both situations where the datasets present unfairness and when they do not. While one-hot encoding can be applied to SVM, our model is a better choice.

The next section dives into numerical tests, demonstrating the efficacy of the proposed methodology.
\section{Simulation Study}
\label{sec:chapter-4}

First we present the step-by-step strategy used to create the datasets and to conduct the numerical experiments. Using \texttt{Julia 1.9} \parencite{BezansonEdelmanKarpinskiShah17} with the packages \texttt{Distributions} \parencite{distributions}, \texttt{DataFrames} \parencite{DataFrames}, \texttt{MLJ} \parencite{MLJ} and \texttt{MKL} \parencite{MKL}. The following parameters are generated randomly. Their specific values will be determined at a later stage.

\begin{itemize}
        \item \emph{Number of points}: Number of points in the dataset; 
        \item \emph{$\beta 's$}: The fixed effects;
        \item \emph{$g's$}: The random effects with distribution $N(0,2)$;
        \item \emph{Data points}: The covariate vector associated with fixed effects with distribution $N(0,1)$;
        \item $c$: Threshold from Fair Support Vector Machine;
        \item \emph{seed}: Random seed used in the generation of data;
        \item \emph{Train-Test split}: Approximately 1\% of the dataset was used for the training set, and 99\% for the test set. This percentage was due to the fact that we randomly selected 3 to 5 points from each cluster for the training set.
\end{itemize}
Then the labels of the synthetic dataset were computed using
\begin{equation}\label{m1}
    m = \beta^T x + g
\end{equation}
in tests where the dataset has random effects, and
\begin{equation}\label{m2}
    m = \beta^T x
\end{equation}
in the tests where the dataset has just fixed effects. Since $m \in [ -\infty, \infty ]$, we project the value to $[ 0, 1 ]$ using the function:
\begin{equation}\label{m21}
    M = \frac{exp(m)}{1+exp(m)}.
\end{equation}
Finally, the true label of each point x is given by:
\begin{equation}\label{m23}
    y = \text{Bernoulli}(M).
\end{equation}
Thus, the datasets are ready.

Regarding the tests, the comparisons are made between the following optimizations problems:
\begin{enumerate}
    \item Mixed Effects Support Vector Machine (MESVM);
    \item Fair Mixed Effects Support Vector Machine (FMESVM);
    \item Support Vector Machine (SVM);
    \item Fair Support Vector Machine (SVMF).
\end{enumerate}
We then, compare the accuracy and the disparate impact of each method above. The test were conducted on a computer with an Intel Core i9-13900HX processor with a clock speed of 5.40 GHz, 64 GB of RAM, and Windows 11 operating system, with 64-bit architecture.

To compute accuracy, first we need compute the classifications. For this, we use expressions \eqref{m1} - \eqref{m23} with
\begin{equation*} \hat{y} =
                \begin{cases}
                  1 \hspace{0.4cm} \text{if } \hspace{0.05cm} M \geq 0.5\\
                  -1 \hspace{0.4cm} \text{if } \hspace{0.05cm} M < 0.5
                \end{cases}\,.
\end{equation*}
Hence given the true label of all points, we can distinguish them into four categories: true positive (TP) or true negative (TN) if the point is classified correctly in the positive or negative class, respectively, and false positive (FP) or false negative (FN) if the point is misclassified in the positive or negative class, respectively. Based on this, we can compute the accuracy, where a higher value indicates a better classification, as follows,
\begin{equation*}
    AC = \dfrac{TP + TN}{TP + TN + FP + FN} \in [0,1].
\end{equation*}

To compute the Disparate Impact of a specific sensitive feature $s$ we use the following equation based on \textcite{DIII},
\begin{equation*}
    \text{di} := \dfrac{\vert\{\ell: \hat{y}_\ell = 1, x_\ell \in \mathcal{S}_0\}\vert}{\vert \mathcal{S}_0 \vert} \dfrac{\vert \mathcal{S}_1 \vert}{\vert\{\ell: \hat{y}_\ell = 1, x_\ell \in \mathcal{S}_1\}\vert} \in [0,\infty).
\end{equation*}
Disparate Impact, as a metric, should ideally be equal to $1$ to indicate fair classifications. Values greater or lower than $1$ suggest the presence of unfairness. For instance, both $\text{di} = 2$ and $\text{di} = 0.5$ represent the same amount of discrimination, but in opposite directions. While the former case deviates further from the ideal value ($1$) compared to the latter, the $\text{di}$ metric itself does not capture this distinction. To address this limitation and achieve a more nuanced metric, we use the minimum value between the $\text{di}$ and its inverse $\tfrac{1}{\text{di}}$ as follows:
\begin{equation}\label{DI}
    \text{DI} := \min (\text{di}, \text{di}^{-1})  \in [0,1].
\end{equation}

The parameters for the Unfair cases are:
\begin{itemize}
     \item $\beta$'s $= [-1.5, 0.4, 0.8, 0.5, 1.5]$;
    \item $g$'s: 100 clusters with $b_i \sim N(0,Q)$, with $i \in [1, 100]$;
    \item $c = 10^{-3}$;
    \item $K = 1$;
    \item $\lambda = 1$.
\end{itemize}
And for the fair cases are:
\begin{itemize}
     \item $\beta$'s $= [-1, 1, 2, 1, 0.1]$;
    \item $g$'s: 100 clusters with $b_i \sim N(0,Q)$, with $i \in [1, 100]$;
    \item $c = 10^{-3}$;
    \item $K = 1$;
    \item $\lambda = 1$,
\end{itemize}
with $Q$ being $2$ for cases with random effect and $0$ for cases without random effect. The $\beta_0$ is related to the intercept, and $\beta_4$ are the coefficient of the binary sensitive feature. In the unfair cases the coefficients were randomly selected using numbers between $0$ and $1$, with the exception of the sensitive feature where we assign a disproportionately high value to the coefficient relative to the other coefficients, thereby emphasizing the greater significance of the sensitive variable in predicting the labels. In other words, data points with the sensitive categories equal to $1$ are more likely to be classified as positive. This practice results in a dataset that is inherently unfair, as needed to test our method. For all experiments, the matrix $X$ was randomly generated from a multivariate normal distribution with zero mean and independent variables. In fair datasets, we substantially decrease the coefficient value of the sensitive feature, aiming to minimize its correlation with the labels.

The following figures present experimental results on four different datasets with 1000 samples each. The datasets are presented individually, and the changes between samples occur only in the training set. Each figure contains four corresponding images representing accuracy and disparate impact for all possible data. All figures were created using the \texttt{Plots}  and \texttt{PlotlyJS} packages, developed by \textcite{Plots}.

\begin{figure}[H]
\centering
\includegraphics[width=12.5cm]{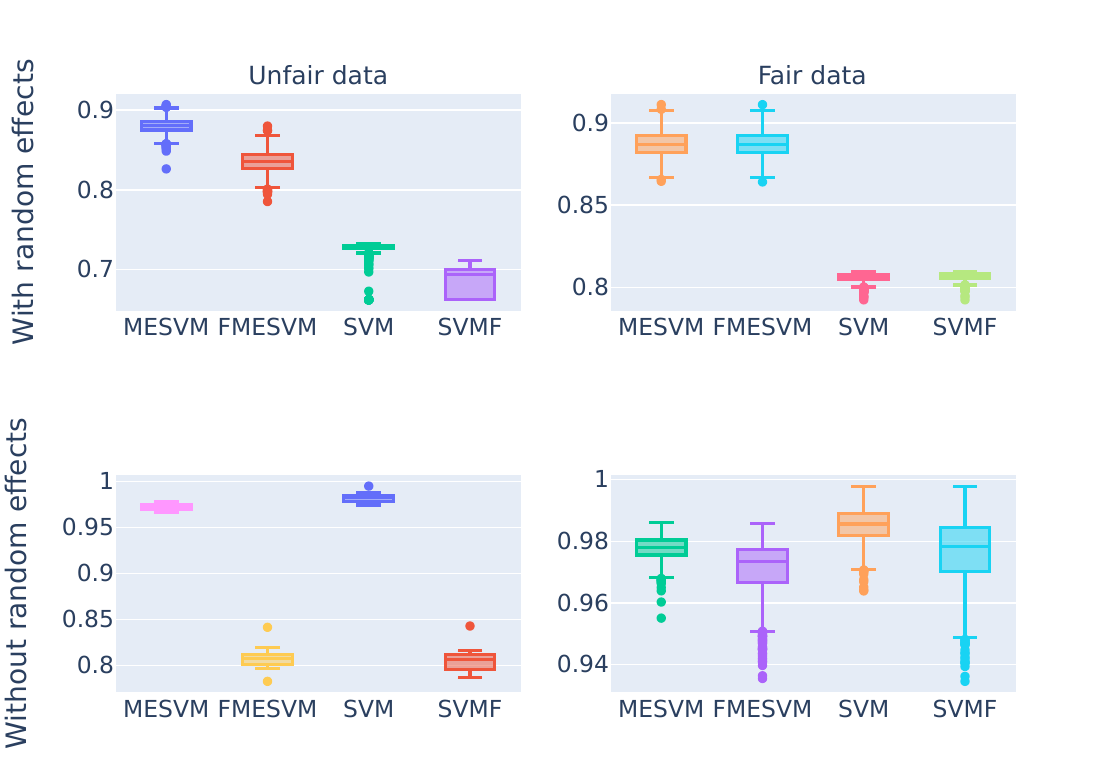}
\caption{Accuracy.}
\label{ACAC}
\end{figure}

\begin{figure}[H]
\centering
\includegraphics[width=12.5cm]{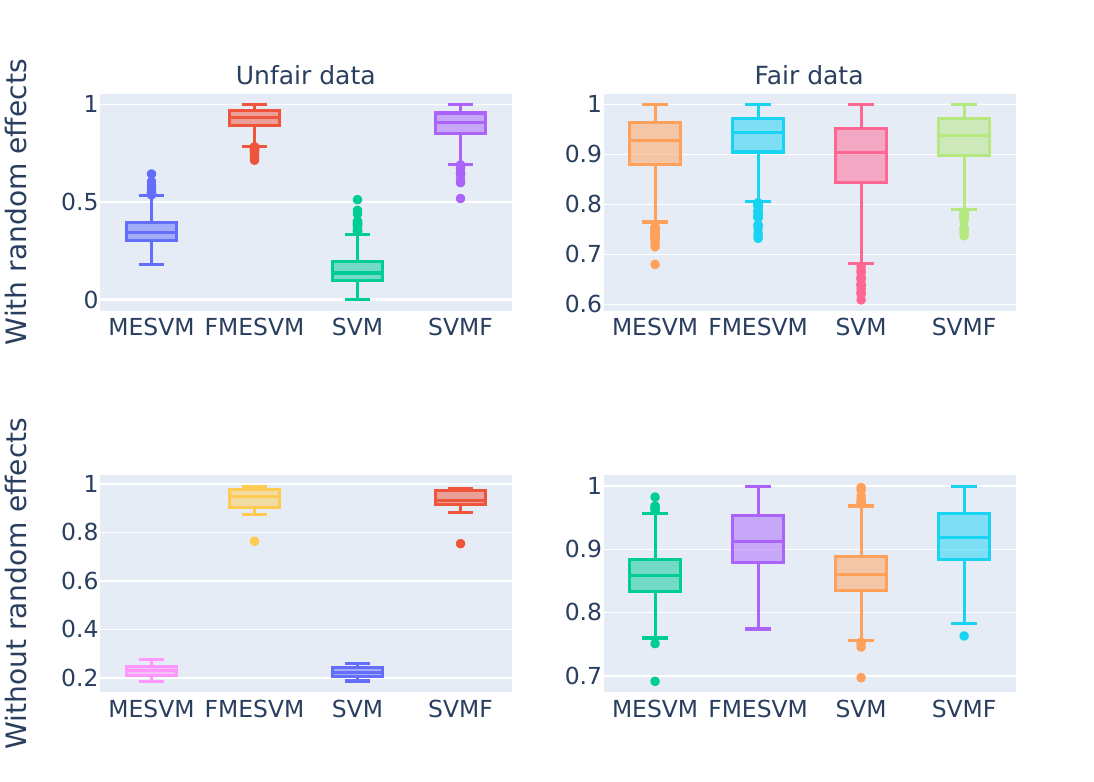}
\caption{Disparate Impact.}
\label{DIDI}
\end{figure}

Figure \ref{ACAC} demonstrates that our proposed approach consistently outperformed alternative methods in scenarios with random effects. In settings without random effects, both methods achieved comparable accuracy, as anticipated. However, a slight decrease in accuracy was observed when introducing unfairness in the data. This is understandable, as the approach must balance addressing both random effects and unfairness simultaneously.

Figure \ref{DIDI} demonstrates that the approach consistently yields improved disparate impact metrics when applied to datasets containing inherent biases. Conversely, in scenarios where the underlying data exhibits no inherent bias, our approach produces equivalent outcomes, as there is no inherent disparity to mitigate.

\section{Application}
\label{sec:chapter5}

In this set of experiments, we do tests using the Adult dataset.  To test the method's efficiency, we created groups based on individuals age and marital status, and a sensitive feature, gender \parencite{speicher2018unified}. The Adult dataset is a famous tool in machine learning, where the goal is predicting whether the individuals earn more ($y = 1$) or less ($y = -1$) than $50000$ USD annually. Were conducted 1000 samples with 0.5\% of the data as the training set and the remaining data as the test set.

The features used in the classification process are the follows:

\begin{multicols}{2}
\begin{itemize}
    \item Age: The age of the individual in years;
    \item Capital\_gain: Capital gain in the previous year;
    \item Capital\_loss: Capital loss in the previous year;
    \item Education: Highest level of education achieved by the individual;
    \item Education\_num: A numeric form of the highest level of education achieved;
    \item Fnlwgt: An estimate of the number of individuals in the population with the same demographics as this individual;
    \item Hours\_per\_week: Hours worked per week;
    \item Marital\_status: The marital status of the individual;
    \item Native\_country: The native country of the individual;
    \item Occupation: The occupation of the individual;
    \item Race: The individual's race;
    \item Relationship: The individual's relationship status;
    \item Gender: The individual's gender;
    \item Workclass: The sector that the individual works in.
\end{itemize}
\end{multicols}

\begin{multicols}{2}
\begin{figure}[H]
\centering
\includegraphics[width=6.15cm]{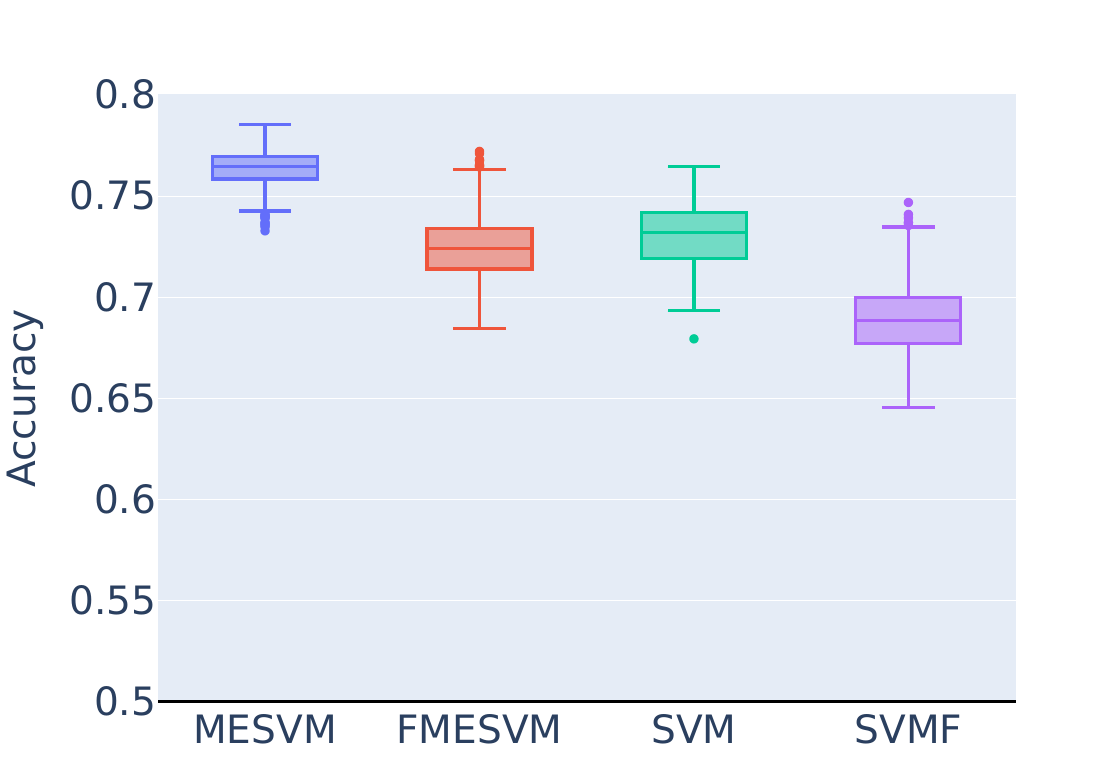}
\caption{Accuracy.}
\label{ACA}
\end{figure}

\begin{figure}[H]
\centering
\includegraphics[width=6.15cm]{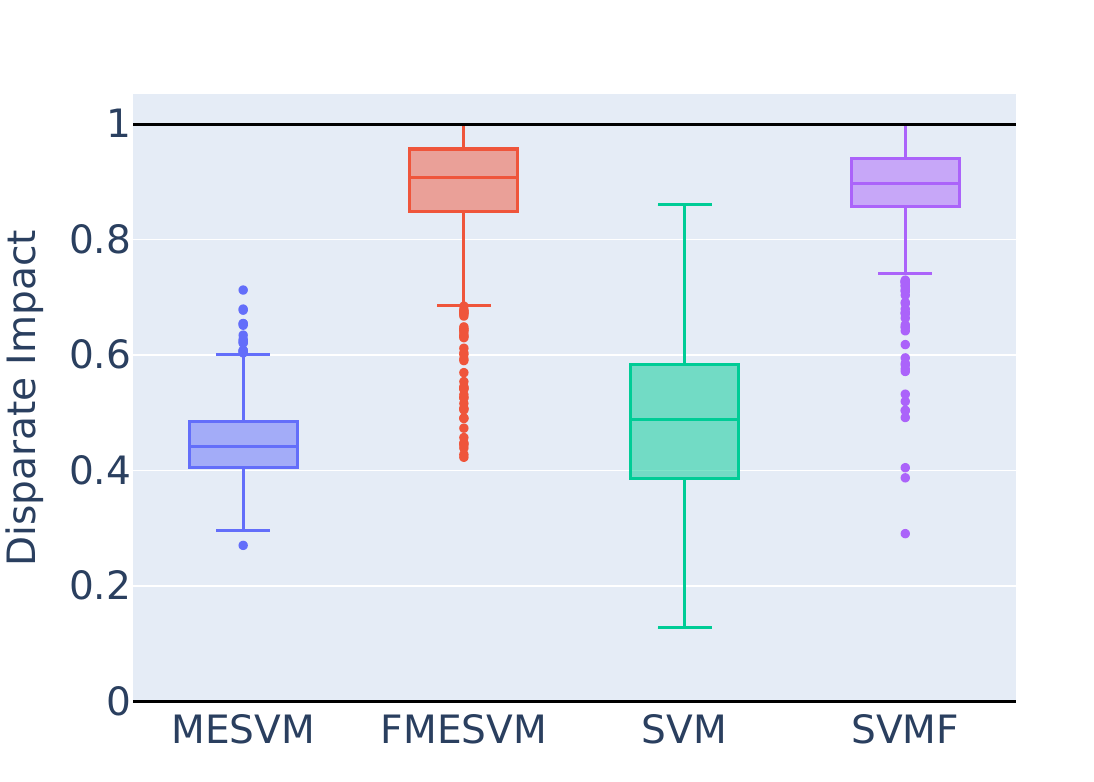}
\caption{Disparate Impact.}
\label{DIA}
\end{figure}
\end{multicols}

As can be seen in Figure \ref{ACA}, in this set of experiments, we obtained a better accuracy in MESVM and FMESVM in comparison to regular SVM since this one not account for random effects.

Figure \ref{DIA} show that we also obtained an improvement in the disparate impact on the Fair algorithms. Note that make sense since we have a unfair population.

Upon examining all the tests, we were able to observe an improvement in Disparate Impact in ($100\%$) of the cases.

\section{Conclusion}
\label{sec:conclusion}

This study investigated a novel approach to mitigating disparate impact, a fairness issue in machine learning models, while simultaneously addressing mixed effects. We introduce a novel Fair Mixed Effects Support Vector Machine (FMESVM) algorithm that tackles both concerns cohesively, overcoming the limitations of existing methods often dedicated to separate problem solving. This integrated approach tailors treatments to the specific demands of each issue, ensuring optimal performance.

Employing the widely respected Support Vector Machine (SVM) for binary classification, the FMESVM framework incorporates mixed effects within the SVM setting and deploys novel regularization techniques to effectively mitigate disparate impact. Extensive evaluation across diverse datasets and metrics demonstrates the success of our proposed method in demonstrably reducing disparate impact while maintaining or minimally compromising overall accuracy.

For comprehensive experimentation, we systematically explored all possible scenarios involving the two concerns: datasets exhibiting both, only unfairness with random effects, only fairness with random effects, and neither issue. This approach yielded expected results, with each combination directly impacting accuracy or disparate impact as predicted.

The FMESVM presents a significant advancement in fairness-aware machine learning by comprehensively addressing disparate impact and mixed effects through a unified framework. This paves the way for the development of more robust and ethical machine learning models with broader applicability.

\section*{Acknowledgements}

The authors are grateful for the support of the German Federal Ministry of Education and Research (BMBF) for this research project, as well as for the \enquote{OptimAgent Project}.

We would also like to express our sincere appreciation for the generous support provided by the German Research Foundation (DFG) within Research Training Group 2126 \enquote{Algorithmic Optimization}.

\printbibliography

\appendix

\end{document}